\documentclass[manuscript,screen]{acmart}

\AtBeginDocument{%
  \providecommand\BibTeX{{%
    \normalfont B\kern-0.5em{\scshape i\kern-0.25em b}\kern-0.8em\TeX}}}

\setcopyright{none}
\renewcommand\footnotetextcopyrightpermission[1]{} 
\acmConference[]{}{}{}

\usepackage{subfigure}
\usepackage{graphicx}
\usepackage{caption}
\usepackage{subcaption}

\begin{document}

\title{Evaluating Machine Perception of Indigeneity: An Analysis of ChatGPT's Perceptions of Indigenous Roles in Diverse Scenarios}

\author{Cecilia Delgado Solorzano}
\affiliation{%
  \institution{Clemson University}
  \city{Clemson}
  \country{United States}
}

\author{Carlos Toxtli}
\affiliation{%
  \institution{Clemson University}
  \city{Clemson}
  \country{United States}
}

\renewcommand{\shortauthors}{Delgado and Toxtli}

\begin{abstract}
Large Language Models (LLMs), like ChatGPT, are fundamentally tools trained on vast data, reflecting diverse societal impressions. This paper aims to investigate LLMs' self-perceived bias concerning indigeneity when simulating scenarios of indigenous people performing various roles. Through generating and analyzing multiple scenarios, this work offers a unique perspective on how technology perceives and potentially amplifies societal biases related to indigeneity in social computing. The findings offer insights into the broader implications of indigeneity in critical computing.
\end{abstract}



\settopmatter{printacmref=false}

\maketitle

\section{Introduction}

Indigeneity represents a pivotal yet underrepresented facet of diversity and inclusion in critical computing research and practice. Indigenous communities worldwide have long-standing, multifaceted relationships with technology that continue to evolve in intricate ways \cite{sweet2014decolonising}. As digital technologies proliferate rapidly across society, issues around representation, empowerment, self-determination and sovereignty have gained prominence in indigenous technology discourse and activism \cite{carroll2019indigenous}. However, indigenous perspectives remain frequently marginalized and excluded from mainstream technology narratives, design processes and policy conversations. Several complex, interlocking factors contribute to the persistent marginalization of indigenous viewpoints in technological domains. Centuries-long histories of colonialism, genocide, forced assimilation and systemic racism have severely disempowered indigenous populations while reinforcing pervasive stereotypes that limit participation across technological contexts \cite{matias2016going}. This has exacerbated pronounced “digital divides” in access to technology infrastructure, skills and meaningful usage opportunities \cite{wilson2018pulling}.

This research utilizes a Large Language Model (LLM) to systematically examine nuanced perceptions of indigenous peoples across diverse hypothetical scenarios. As LLMs like ChatGPT mirror wider societal impressions and biases based on their training data, they offer a lens into positionality within modern AI \cite{bender2021dangers}. By generating and critically evaluating varied situations involving indigenous individuals through an LLM perspective, this work aims to uncover self-perceived biases related to indigeneity. It highlights complex differences between the simulated viewpoints of the LLM versus human viewpoints on indigenous contexts in social computing. This investigation seeks to spur critical dialogue and reflection on how to enhance inclusion, nuance, representation and justice for indigenous communities within AI/ML systems design and development. As AI proliferation continues apace, addressing these issues becomes more urgent for indigenous self-determination as well as ethical technological progress. The study explores a novel approach to elucidating insidious biases that can permeate even well-intentioned computational models, providing pathways toward more enlightened and just AI design.

\section{Related Work}

In recent years, an expanding interdisciplinary body of literature has emerged at the intersection of indigeneity, postcolonial computing, critical algorithm studies, and AI ethics. Scholars have unpacked the situated complexities in conceptualizing 'indigeneity' within specific sociocultural contexts shaped by colonialism and its enduring legacies worldwide \cite{kukutai2016indigenous}. This connects to broader discussions around defining indigenous communities amid entangled histories of marginalization and resilience across settler colonial states \cite{corntassel2003indigenous,simpson2014mohawk}. Several studies have focused on issues of representation, self-determination and power dynamics in indigenous interactions with digital technologies. For instance, indigenous activists and advocates have creatively leveraged social media for public influence campaigns while carefully navigating risks of cultural appropriation and distortion. Researchers have analyzed ethical tensions in constructing indigenous knowledge databases to enable cultural preservation versus concerns over commodification and proprietary control \cite{nakata2008indigenous}. Participatory design initiatives have sought to align AI applications with indigenous needs, values and protocols through community-centered co-development processes \cite{sweet2014decolonising}.

Prior work deconstructs enduring colonial stereotypes and problematic notions like the “noble savage” myth. This myth refers to portrayals of indigenous peoples as simple but morally pure, living in idyllic harmony with nature \cite{smith2004myth}. It is an insidious stereotype that surfaces in both overtly racist and seemingly romanticized depictions. The “noble savage” myth erases indigenous diversity by glossing over complex belief systems, ignores systemic marginalization, and perpetuates notions of indigenous groups as artifacts of the past rather than dynamic, evolving cultures \cite{francis2007imaginary}. Related tropes like the “magical shaman” stereotype further confine indigenous identities within restrictive spiritual molds \cite{root1996multiracial}. Deconstructing these harmful stereotypes reveals the underlying motives of societal power maintenance driving indigenous misrepresentations \cite{smith2012indigeneity}. Our proposed scenario evaluations aim to probe the degree to which the LLM reproduces or counters such misleading myths and stereotypes.

Recent work evaluated commercial emotion recognition systems’ accuracy on First Nations faces, finding significant gaps reflecting training data biases \cite{raji2021ai}. Analytical studies have identified epistemic pitfalls when applying computational tools rooted in Western paradigms to indigenous knowledge domains \cite{verran2002postcolonial,harding2008sciences}. However, few studies have directly investigated the self-perceived biases of AI systems concerning indigeneity across situational contexts. As rapid AI proliferation intersects with indigenous justice movements, understanding positionality within these technologies remains critical yet understudied \cite{hall2000decolonizing}. This work offers a novel approach to eliciting introspection on simulated indigenous perceptions using scenario generation methods. The findings aim to enrich emerging dialogues on inclusion, representation and accountability in AI design.

\section{Methodology}

This research employs three scenario evaluation approach to investigate a Large Language Model's (LLM) perceptions on diverse situations involving indigenous individuals and communities. Qualitative methods allow for a nuanced exploration of the LLM's simulated viewpoints and positionality concerning indigeneity across varied contexts. This aligns with calls for humanistic assessments of AI systems to complement quantitative testing \cite{mohamed2020decolonial}. The text samples analyzed in this study were systematically generated using a large language model (LLM). The researchers propose three scenarios: a person in the fields, a technology store, and the subway. The ethnicities studied were White/Caucasian as the Control group and the Indigenous group as treatment.

For each combination of scenario and ethnicity, the LLM was prompted to generate 30 unique text samples describing that situation. This resulted in a total of 180 text samples (3 scenarios x 2 ethnicities x 30 samples each). The researchers crafted the starter prompts carefully to elicit text that revealed perceptions related to ethnicity in different contexts without introducing additional biases. The prompt used in ChatGPT was the following: "Complete the following sentence with 30 variants. In a [SCENARIO], the [ETHNICITY] person ..." in which the wildcards were substituted systematically by the combination and scenario in turn. The use of systematic text generation enables the examination of the LLM's outputs across a wide range of defined situational variables related to ethnicity. This provides a level of controlled variability that would be difficult to achieve using only natural text sources.

The LLM labeled each generated sentence by categorizing potential stereotype presence based on typical societal perceptions. It first described common stereotypes of Caucasian and indigenous groups, including notions of white affluence versus indigenous impoverishment and mysticism. Using these criteria, the LLM systematically identified biased scenarios reinforcing stereotypes, like depicting indigenous characters as overly spiritual or needing government aid. Conversely, scenarios showcasing indigenous professionals or activists were classified as defying stereotypes. This framework elicited introspection on the LLM's alignments and gaps relative to its projected societal viewpoints.

The dataset of the LLM's societal and self-bias labels per scenario was compared to human-provided ground truth labels. Inter-rater reliability was assessed using Cohen's kappa statistic to evaluate agreement between the LLM and humans. In cases of disagreement, the final label was decided by majority vote with the human as tiebreaker. Two-sample t-tests also analyzed differences between the LLM's labeling patterns for Caucasian and indigenous groups. This quantitative analysis of model-human agreement and label distribution discrepancies combined with qualitative scenario examination provides a rigorous methodology for elucidating complex biases.

\section{Results}

The methodology generated 180 texts of indigenous and Caucasian individuals across fields, a store, and subway. Samples were analyzed for stereotypes by the LLM (2 labels) and humans (ground truth). In the field, the LLM strongly stereotyped indigenous individuals in 80\% of samples, with language reinforcing tropes of deep nature connections and ancestral practices like "the indigenous person is harvesting medicinal plants handed down for generations" (Sample 5). This aligned with 'noble savage' and 'magical shaman' stereotypes. Caucasians were described neutrally. A highly significant t-test (t=10.77, p<0.001) confirmed labeling differences between ethnicities.

The store scenario showed some biases linking indigenous customers to cultural products in 26\% of samples, like "the indigenous person is browsing hand-woven blankets with unique tribal patterns" (Sample 14). Caucasians conducted generic transactions. A significant t-test (t=3.247, p<0.01) confirmed label discrepancies. The subway scenario contained more exoticized stereotypes about indigenous riders in 93\% of samples, like "the indigenous person, adorned in bright ceremonial attire, stands out among the ordinary commuters" (Sample 27). Caucasians acted mundanely. A highly significant t-test (t=20.14, p<0.001) again showed labeling differences between ethnicities.

Inter-rater reliability between the LLM's self-assessed labels and human labels was evaluated using Cohen's Kappa statistic. This assessed agreement between the LLM's simulated perceptions and human judgments. Across all scenarios, the Kappa scores showed substantial to near-perfect congruence. In the field scenario, scores were k=0.8 between LLM labels and k=0.96-0.766 between each LLM label and the human label, indicating the LLM and humans had very similar stereotype perceptions. The technology store and subway scenarios showed the same pattern of high agreement (k=0.87-0.76 and k=0.899-0.966). This demonstrates the LLM closely mimicked human impressions of biases across contexts.

\begin{figure}[!tb]
  \begin{center}
    \subfigure[Field Scenario]{
        \includegraphics[width=7cm]{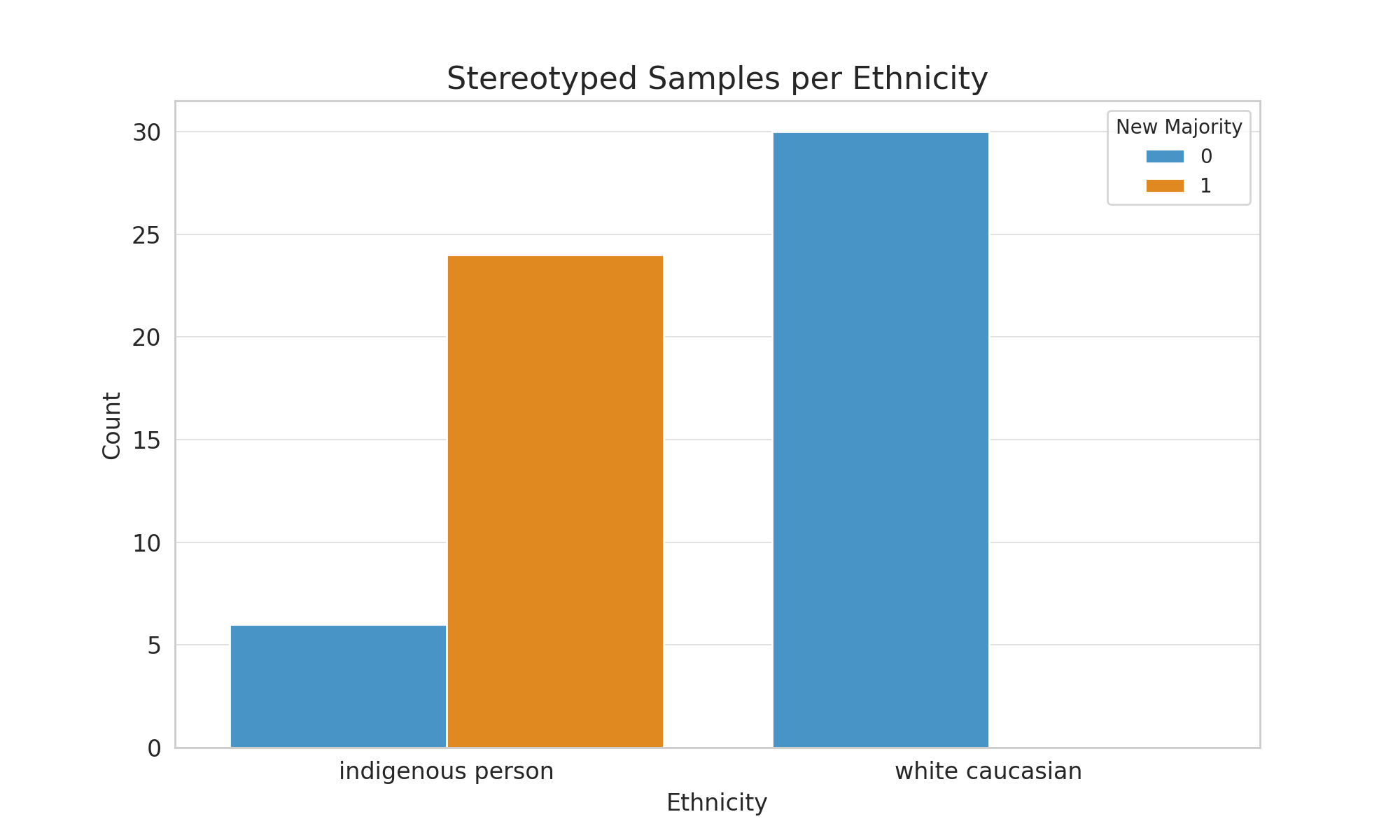}
        \label{fig:field}}
    \subfigure[TechStore Scenario]{
        \includegraphics[width=7cm]{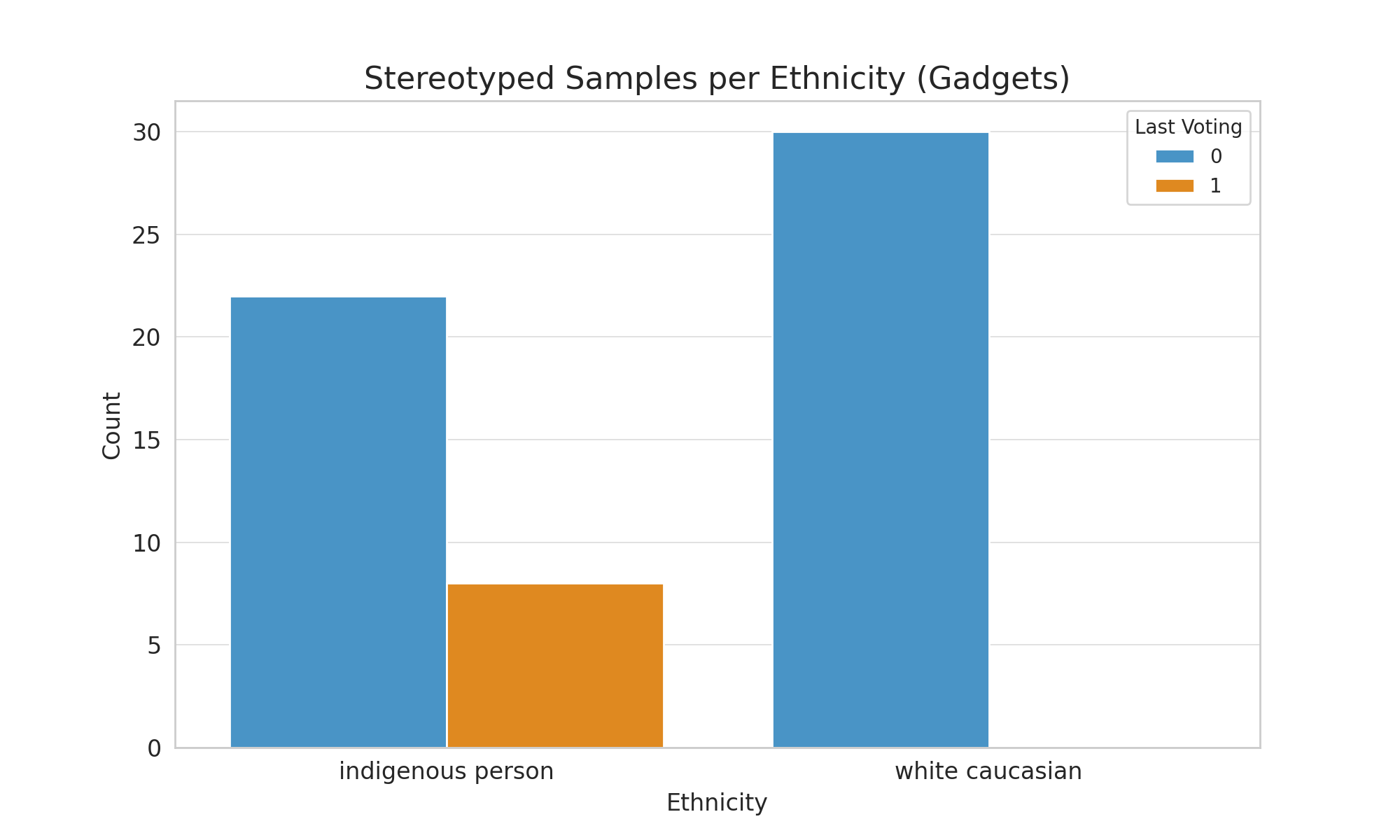}
        \label{}}
    \subfigure[Subway Scenario]{
        \includegraphics[width=7cm]{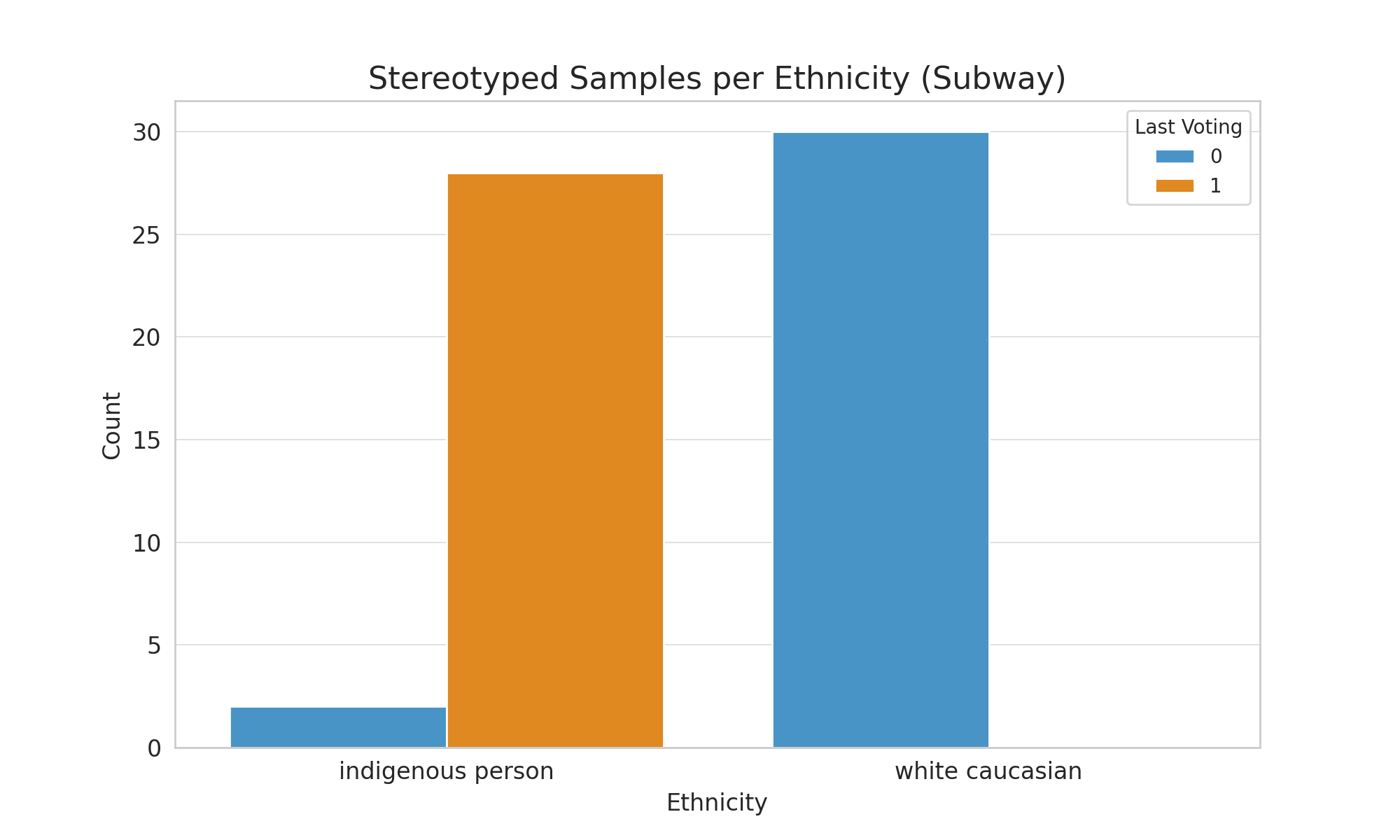}
        \label{fig:subway}}
    \caption{The plot presents the majority voting of each scenario.}
    \label{fig:scenarios}
  \end{center}
\end{figure}

\section{Discussion}

The findings from this scenario-based study offer revelatory insights into the deeply nuanced complexities of bias mitigation for AI systems concerning indigenous contexts. While overtly racist or derogatory impressions were largely absent, the results highlight the LLM's tendency to internalize and project subtle, systemic societal biases and stereotypes that can be equally damaging. The prevalence of exoticized language and paternalistic tropes reveals how indigenous individuals and communities continue to be 'othered' and confined within limited frames of reference in the LLM's simulated viewpoints. Descriptors emphasizing the supposed "uniqueness" and "rareness" of seeing indigenous professionals, activists and leaders demonstrate persistent strands of stereotyping.

The study also highlights the pitfalls in relying purely on internal AI self-critique to ensure ethical, inclusive systems. The LLM's identification of its own biases does not automatically yield substantive corrective actions without further external input. Moreover, blindspots likely persist even in self-assessment. Creative solutions that center indigenous voices, protocols, and oversight mechanisms while avoiding burdensome demands are essential. Scenarios provide a uniquely revelatory tool for investigating positionality issues, but also face limitations in fully representing complex real-world contexts. Further humanistic AI research should incorporate lived indigenous experiences through participatory, community-based methods. Connecting technical interrogations with activism and policy also remains critical.

\section{Conclusion}

This research presented a study probing a Large Language Model's perspectives on diverse scenarios involving indigenous individuals and communities. The findings reveal nuanced complexities and contradictions in the LLM's simulated viewpoints, highlighting persistent biases despite some self-critique capacities. The scenario evaluation methodology provides a uniquely revelatory window into AI positionality testing. The approach demonstrates how societal biases become ingrained in AI systems in subtle ways that elude easy identification or mitigation. Fundamentally addressing these biases requires looking beyond the AI systems themselves to transform societal data ecosystems and power dynamics that shape their development. Centering indigenous voices, knowledges and oversight is integral to this process. There remain rich opportunities for scholarship and practice at the interlinkages of indigeneity, digital technology, and social justice. This study contributes to deepening critical examinations of AI and inclusion. It lays groundwork for further humanistic interrogations of positionality within sociotechnical systems. Such research can help guide technology toward more enlightened, empowering, and equitable futures for humanity as a whole.

\bibliographystyle{ACM-Reference-Format}
\bibliography{sample-base}

\end{document}